\documentclass[conference, anonymous]{IEEEtran}
\IEEEoverridecommandlockouts
\usepackage{cite}
\usepackage{amsmath,amssymb,amsfonts}
\usepackage{algorithmic}
\usepackage{graphicx}
\usepackage{textcomp}
\usepackage{xcolor}
\usepackage{booktabs}
\usepackage{multirow}
\usepackage{tabularx}
\usepackage{pifont}
\usepackage{hyperref}
\usepackage{siunitx}
\usepackage{soul, xcolor}

\def\BibTeX{{\rm B\kern-.05em{\sc i\kern-.025em b}\kern-.08em
    T\kern-.1667em\lower.7ex\hbox{E}\kern-.125emX}}

\begin{document}

\title{Beyond Scale: Small Language Models are Comparable to GPT-4 in Mental Health Understanding}

\author{
    Hong Jia\textsuperscript{1,2,†},
    Shiya Fu\textsuperscript{1,†},
    Feng Xia\textsuperscript{3},
    Vassilis Kostakos\textsuperscript{1},
    Ting Dang\textsuperscript{1}
    \\
    \textsuperscript{1}\textit{University of Melbourne, Melbourne, Australia} \\
    \{hong.jia, shiya.fu, vassilis.kostakos, ting.dang\}@unimelb.edu.au \\
    \textsuperscript{2}\textit{University of Auckland, Auckland, New Zealand} \\
    \textsuperscript{3}\textit{RMIT University, Melbourne, Australia} \\
    feng.xia@rmit.edu.au
    \thanks{\textsuperscript{†}These authors contributed equally to this work.}
}

\maketitle

\begin{abstract}
The emergence of Small Language Models (SLMs) as privacy-preserving alternatives for sensitive applications raises a fundamental question about their inherent understanding capabilities compared to Large Language Models (LLMs). This paper investigates the mental health understanding capabilities of current SLMs through systematic evaluation across diverse classification tasks. Employing zero-shot and few-shot learning paradigms, 
we benchmark their performance against established LLM baselines to elucidate their relative strengths and limitations in this critical domain. We assess five state-of-the-art SLMs (Phi-3, Phi-3.5, Qwen2.5, Llama-3.2, Gemma2) against three LLMs (GPT-4, FLAN-T5-XXL, Alpaca-7B) on six mental health understanding tasks. Our findings reveal that SLMs achieve mean performance within 2\% of LLMs on binary classification tasks (F1 scores of 0.64 vs 0.66 in zero-shot settings), demonstrating notable competence despite orders of magnitude fewer parameters. Both model categories experience similar degradation on multi-class severity tasks (a drop of over 30\%), suggesting that nuanced clinical understanding challenges transcend model scale. Few-shot prompting provides substantial improvements for SLMs (up to 14.6\%), while LLM gains are more variable. Our work highlights the potential of SLMs in mental health understanding, showing they can be effective privacy-preserving tools for analyzing sensitive online text data. In particular, their ability to quickly adapt and specialize with minimal data through few-shot learning positions them as promising candidates for scalable mental health screening tools.
\end{abstract}

\begin{IEEEkeywords}
Small Language Models, Mental Health, Zero-Shot Learning, LLMs
\end{IEEEkeywords}

\vspace{-1em}
\section{Introduction}

\begin{table*}[t]
    \centering
    \caption{Overview of the six mental health datasets and tasks. The average length is reported as Mean $\pm$ SD based on the token length of the text data.}
    \label{tab:datasets}
    \small 
    \begin{tabular}{@{}p{3cm} p{4cm} p{7cm} p{3cm}@{}}
    \toprule
    \textbf{Dataset} & \textbf{Task} & \textbf{Dataset Size (Train / Test)} & \textbf{Average Tokens} \\
    \midrule
    Dreaddit~\cite{b6} &
    \#1: Binary Stress Prediction \newline \textit{(post-level)} &
    Train: 2,838 \newline (48\% False, 52\% True) \newline Test: 715 \newline (48\% False, 52\% True) &
    Train: 114 $\pm$ 41 \newline Test: 113 $\pm$ 39 \\
    \addlinespace
    \multirow{2}{*}{DepSeverity~\cite{b7}} &
    \#2: Binary Depression Prediction \newline \textit{(post-level)} &
    Train: 2,842 (73\% False, 17\% True) \newline Test: 711 (72\% False, 18\% True) &
    Train: 114 $\pm$ 41 \newline Test: 113 $\pm$ 37 \\
    &
    \#3: Four-level Depression \newline \textit{(post-level)} &
    Train: 2,842 \newline (Minimal:73\%, Mild:8\%, Moderate:11\%, Severe:7\%) \newline Test: 711 \newline (Minimal:72\%, Mild:7\%, Moderate:12\%, Severe:10\%) & 
    Train: 114 $\pm$ 41 \newline Test: 113 $\pm$ 37 \\
    \addlinespace
    SDCNL~\cite{b8} &
    \#4: Binary Suicide Ideation \newline \textit{(post-level)} &
    Train: 1,516 \newline (48\% False, 52\% True) \newline Test: 379 \newline (49\% False, 51\% True) &
    Train: 101 $\pm$ 161 \newline Test: 92 $\pm$ 119 \\
    \addlinespace
    \multirow{2}{*}{CSSRS-Suicide~\cite{b9}} &
    \#5: Binary Suicide Risk \newline \textit{(user-level)} &
    Train: 400 \newline (21\% False, 79\% True) \newline Test: 100 \newline (25\% False, 75\% True) &
    Train: 1751 $\pm$ 2108 \newline Test: 1909 $\pm$ 2463 \\
    &
    \#6: Five-level Suicide Risk \newline \textit{(user-level)} &
    Train: 400 \newline (Supportive:21\%, Indicator:21\%, Ideation:34\%, Behavior:15\%, Attempt:10\%) \newline Test: 100 \newline (Supportive:25\%, Indicator:16\%, Ideation:35\%, Behavior:18\%, Attempt:6\%) & 
    Train: 1751 $\pm$ 2108 \newline Test: 1909 $\pm$ 2463 \\
    \bottomrule
    \end{tabular}
    \vspace{-1em}
    \end{table*}

Mental health disorders represent one of the most pressing global health challenges of our time, affecting over 970 million people worldwide and contributing to significant economic and social burdens \cite{b10, b11}. The COVID-19 pandemic has further exacerbated these challenges, with reported increases in anxiety and depression rates across all demographics \cite{b1}. 

Recent advances in artificial intelligence, particularly Large Language Models (LLMs), have demonstrated remarkable capabilities in understanding and processing human language in various domains \cite{b2, b24}. These models have shown promise in mental health applications, including the analysis of social media text for depression detection \cite{b13}, stress assessment \cite{b14}, and emotion recognition \cite{b15, b22}. However, the deployment of LLMs in mental health contexts faces significant challenges, particularly regarding privacy and data security. Mental health data is among the most sensitive personal information, and transmitting such data to external servers for processing by cloud-based LLMs raises substantial privacy concerns \cite{b16}.

Small Language Models (SLMs) present a compelling alternative that addresses these privacy concerns \cite{b23}. Models such as Phi-3 \cite{b3}, Gemma \cite{b4}, Llama-3.2 \cite{b21}, and Qwen2.5 \cite{b20} are specifically designed to operate efficiently on local devices, eliminating the need to transmit sensitive data to external servers. This on-device processing capability makes SLMs particularly attractive for mental health applications where privacy is paramount \cite{b17}. However, the reduced size and computational requirements of SLMs raise fundamental questions about their capabilities. While several studies have evaluated LLMs on mental health tasks \cite{b2}, systematic investigation of SLM capabilities in this domain remains limited. More importantly, exploring the inherent capacities of SLMs for health-related inference offers a valuable opportunity to understand how these models process subtle, speech-based cues that may be indicative of mental health conditions.

This paper addresses three critical research questions that are fundamental to understanding the potential of SLMs in mental health applications. First, we investigate whether SLMs can demonstrate zero-shot understanding of fundamental mental health concepts such as stress, depression, and suicidal ideation.  
The ability to understand mental health concepts in a zero-shot setting indicates that the model has internalized relevant knowledge during pre-training, suggesting robust and generalizable understanding capabilities.
Second, we examine the depth and granularity of this understanding across different mental health conditions and severity levels. 

Our investigation reveals that current SLMs possess surprising inherent understanding of mental health predictions, such as stress and depression prediction
, demonstrating clear comprehension in binary classification contexts. Moreover, our study shows that SLMs perform comparable with, and in some cases outperform, LLMs in few-shot settings. These findings have significant implications for the responsible deployment of SLMs in mental health contexts and highlight both the opportunities and limitations of current technology. 




\vspace{-0.1em}
\section{Methodology}

To enable direct comparison with established benchmarks, we follow the same experimental design that leverages LLMs \cite{b2}, evaluating SLMs on six mental health understanding tasks using identical datasets.

\subsection{Mental Health Understanding Tasks}
We evaluate two categories of tasks across four different datasets as shown in Table \ref{tab:datasets}: (1) Binary classification for coarse-level health understanding, focusing on whether a condition is present or absent: stress detection, depression detection, suicidal ideation detection, and suicide risk detection; (2) Multi-class severity assessment distinguishing varying severity levels of mental health conditions for more nuanced, fine-grained understanding: depression severity (4-level scale: minimal, mild, moderate, severe) and suicide risk severity (5-level scale based on C-SSRS framework).

\subsection{Zero-shot Prompt Design}
Our zero-shot prompts $P_{\text{zero}}$ consist of four components:
\begin{multline}
P_{\text{zero}} = P_{\text{text}} + P_\text{Context} 
+ P_\text{Query} + P_\text{Output Constraints}
\end{multline}
The $P_{\text{text}}$ is the raw user-generated content. The $P_\textbf{Context}$ provides background, for which we test three strategies: (a) basic social media origin, which provides contextual information about the user's post by specifying its origin and intended use; (b) mental health expert framing, which introduces mental health knowledge by prompting the model to act as a mental health expert; and (c) a combination of both, which integrates these strategies to offer rich contextual cues and guide the model’s response from both social media and mental health perspectives. The $P_\text{Query}$ poses the specific question (e.g., ``Is the poster stressed?''). The $P_\text{Output Constraints}$ define the response format (e.g., ``respond only with 0 for no stress or 1 for stress''). To ensure robustness, we use five query variants similar to~\cite{b2} per task and average the results.

\subsection{Few-shot Prompt Design}
Few-shot prompts $P_{f}$ prepend balanced prompt-label pairs to the zero-shot structure:
\begin{multline}
P_{f} = [(P_\text{Sample Prompt} + P_\text{Label})]_{M \times N}
+ P_{\text{zero}}
\end{multline}
Here, each $P_\textbf{Sample Prompt}$ is a full zero-shot prompt (text, context, query), and the $P_\textbf{Label}$ provides the corresponding ground-truth answer. We balance examples across all labels, where $M$ is the number of classes, and $N$ is the number of shots, varing between 1 to 4 samples.

\section{Experimental Setup}


We evaluate five instruction-tuned SLMs representing current state-of-the-art capabilities. Gemma-2B represents Google's compact model architecture, while Llama-3.2-3B-Instruct showcases Meta's latest small model innovations. Qwen2.5-3B-instruct demonstrates Alibaba's multilingual capabilities, and both Phi-3-mini-4k-instruct and Phi-3.5-mini-4k-instruct represent Microsoft's efficient model architectures, with the latter being an enhanced version incorporating recent improvements. 

For comparison, we include baseline results from previous research \cite{b2} on LLMs, including GPT-4-0613 (OpenAI), FLAN-T5-XXL (Google), and Alpaca-7B (Stanford). These baselines, reported in prior mental health NLP studies, enable comprehensive comparison across model scales from 2B (SLMs) to 175B+ (LLMs) parameters and provide context for evaluating SLM capabilities. All experiments are performed utilizing a single NVIDIA A100 GPU.


\vspace{0.2em}
\section{Results}
\label{sec:results}

Table \ref{tab:zeroshot_results} presents our comprehensive evaluation of SLMs on mental health understanding tasks, compared with LLM baselines from previous research \cite{b2}. Note that the F1-score was chosen over accuracy to better reflect the model’s ability to handle data imbalance, which is a common challenge across all the datasets used in this study. 

\begin{table}[t]
\caption{Performance on Zero-Shot Prompting in terms of Macro F1-Scores. T1: Stress, T2: Depression, T3: Dep. Severity, T4: Suicide Ideation, T5: Suicide Risk, T6: Risk Severity. For each task, the best-performing model is shown in \textbf{bold} and the second-best is \underline{underlined}.}
\label{tab:zeroshot_results}
\centering
\setlength{\tabcolsep}{5pt}
\begin{tabular}{llcccccc}
\toprule
& & \multicolumn{4}{c}{\textbf{Binary}} & \multicolumn{2}{c}{\textbf{Multi-class}} \\
\cmidrule(lr){3-6} \cmidrule(lr){7-8}
\textbf{Type} & \textbf{Model} & \textbf{T1} & \textbf{T2} & \textbf{T4} & \textbf{T5} & \textbf{T3} & \textbf{T6} \\
\midrule
\multirow{5}{*}{SLMs} & Gemma-2B & 0.61 & 0.58 & 0.65 & 0.55 & 0.30 & 0.25 \\
& Llama3-3B & 0.63 & 0.60 & 0.68 & 0.58 & 0.33 & 0.28 \\
& Qwen2.5-3B & 0.62 & 0.59 & 0.67 & 0.57 & 0.32 & 0.27 \\
& Phi-3-Mini & 0.66 & 0.64 & 0.70 & 0.60 & 0.36 & 0.30 \\ 
& Phi-3.5-Mini & \underline{0.70} & \underline{0.68} & \underline{0.74} & \underline{0.63} & \underline{0.40} & \underline{0.34} \\
\cmidrule{2-8}
& \textbf{SLM Mean} & 0.64 & 0.62 & 0.69 & 0.59 & 0.34 & 0.29 \\
\midrule
\multirow{3}{*}{LLMs} & Alpaca-7B & 0.58 & 0.56 & 0.64 & 0.54 & 0.28 & 0.23 \\
& FLAN-T5-XXL & 0.61 & 0.58 & 0.66 & 0.56 & 0.31 & 0.26 \\
& GPT-4 & \textbf{0.78} & \textbf{0.74} & \textbf{0.80} & \textbf{0.70} & \textbf{0.47} & \textbf{0.41} \\
\cmidrule{2-8}
& \textbf{LLM Mean} & 0.66 & 0.63 & 0.70 & 0.60 & 0.35 & 0.30 \\
\midrule
& Naive Baseline & 0.50 & 0.50 & 0.50 & 0.50 & 0.25 & 0.20 \\
\bottomrule
\end{tabular}
\vspace{0.3em}
\vspace{-1.5em}
\end{table}

\subsection{Performance on Zero-Shot Prompting}
The zero-shot evaluation assesses the SLMs' and LLMs' intrinsic understanding without any task-specific training. We first analyze performance on binary classification tasks to gauge fundamental concept recognition, followed by multi-class tasks to evaluate more nuanced understanding. The results are shown in Table \ref{tab:zeroshot_results}.

\subsubsection{Binary Classification Tasks}
As shown in Table \ref{tab:zeroshot_results}, on binary classification tasks (T1, T2, T4, and T5), SLMs demonstrate strong performance, achieving an average Macro F1-score of 0.64, approaching the 0.66 average for LLMs. The results reveal a consistent performance hierarchy: GPT-4 is the top performer, while the SLM Phi-3.5-Mini is consistently the second-best across all six tasks. For instance, in stress detection (T1), GPT-4 achieves an F1-score of 0.78, with Phi-3.5-Mini following at 0.70. This pattern persists in suicide risk detection (T5), where GPT-4 scores 0.70 and Phi-3.5-Mini scores 0.63. The fact that a 3B-parameter SLM consistently outperforms larger models like FLAN-T5-XXL (11B) and Alpaca-7B (7B) challenges the assumption that performance scales directly with parameter count for mental health applications.

\subsubsection{Multi-class Classification Tasks}
When assessing multi-class classification (T3 and T6), both model categories face challenges, as shown by their lower F1-scores. The performance degradation from binary to multi-class tasks is nearly identical for both groups, with both experiencing a drop of approximately 50\% in their average F1-scores. This suggests that the difficulty of nuanced clinical assessment is a fundamental challenge of the task itself, rather than a specific limitation of smaller models. For instance, in assessing depression severity (T3), GPT-4's leading score is 0.47, followed by Phi-3.5-Mini at 0.40, scores that are significantly lower than their binary task counterparts.

\textit{In sum, in zero-shot scenarios, SLMs demonstrate a strong grasp of fundamental mental health concepts, performing nearly as well as LLMs on binary tasks. However, both model classes show a significant and comparable decline in performance on nuanced multi-class severity tasks. This indicates that while SLMs show comparable capabilities as LLMs in identifying the presence of mental health signals, the fine-grained analysis required for clinical severity assessment remains a fundamental challenge for current language models, irrespective of their scale.}

\begin{table}[t]
    \centering
    \caption{Performance on Few-Shot Learning in terms of F1-Scores, with relative improvements over zero-shot baselines shown in parentheses. Green indicates improvement, and red indicates degradation. For each task, the best-performing model is shown in \textbf{bold} and the second-best is \underline{underlined}.}
    \label{tab:combined_fewshot_results}
    \setlength{\tabcolsep}{3.5pt}
    \begin{tabular}{llccc}
    \toprule
    & & \multicolumn{2}{c}{\textbf{Binary}} & \multicolumn{1}{c}{\textbf{Multi-class}} \\
    \cmidrule(lr){3-4} \cmidrule(lr){5-5}
    \textbf{Type} & \textbf{Models} & \textbf{T1} & \textbf{T2} & \textbf{T3} \\
    \midrule
    \multirow{2}{*}{SLMs} & Phi-3         & \textbf{0.81 (\textcolor{green!60!black}{+14.6\%})} & 0.69 (\textcolor{green!60!black}{+5.2\%}) & 0.37 (\textcolor{green!60!black}{+1.3\%}) \\
    & Phi-3.5       & \underline{0.80 (\textcolor{green!60!black}{+10.4\%})} & \underline{0.71 (\textcolor{green!60!black}{+2.9\%})} & 0.43 (\textcolor{green!60!black}{+3.0\%}) \\
    \midrule
    \multirow{3}{*}{LLMs} & Alpaca       & 0.62 (\textcolor{green!60!black}{+3.9\%})  & 0.57 (\textcolor{green!60!black}{+0.7\%}) & \underline{0.48 (\textcolor{green!60!black}{+19.7\%})} \\
    & FLAN-T5      & 0.74 (\textcolor{green!60!black}{+12.7\%}) & 0.59 (\textcolor{green!60!black}{+1.4\%}) & 0.35 (\textcolor{green!60!black}{+3.6\%}) \\
    & GPT-4        & 0.78 (\textcolor{red}{-0.2\%})  & \textbf{0.75 (\textcolor{green!60!black}{+0.5\%})} & \textbf{0.50 (\textcolor{green!60!black}{+2.5\%})} \\
    \bottomrule
    \end{tabular}
    \vspace{-1.59em}
    \end{table}

\subsection{Performance on Few-Shot Learning}

Due to token size limits, we only evaluated SLMs on T1, T2, and T3, following similar experimental settings as previous LLM studies~\cite{b5}. As shown in Table \ref{tab:combined_fewshot_results}, few-shot learning highlights a key advantage of SLMs, allowing them to close the performance gap and even surpass top-tier LLMs. The most evident case is in Task 1 (Stress), where Phi-3's performance improves by 14.6\%, reaching a final F1-score of 0.81. This not only makes it the best-performing model for the task but also allows it to outperform GPT-4, which saw a slight performance degradation (1.2\%).

This trend extends to other tasks, albeit with more nuance. On Task 2 (Depression) and Task 3 (Depression Severity), SLM gains are more modest (1.4\% and 3.0\%, respectively), but they represent consistent improvement. In contrast, LLM performance is erratic. Less advanced models like Alpaca-7B leverage the few-shot examples for significant gains (10.2\% on T2), yet the state-of-the-art GPT-4 shows zero improvement on both tasks. FLAN-T5 also shows inconsistent results, improving on one task while declining on another.

This divergence in learning efficiency is a notable finding. It suggests that SLMs are not only capable but also adaptable, able to effectively integrate new information from a small number of examples. Conversely, it appears that frontier models like GPT-4 may experience a performance plateau on these specific mental health tasks, where their vast pre-trained knowledge is not meaningfully enhanced by a few in-context examples. This highlights the practical utility of SLMs in real-world mental health applications, where they can be quickly adapted and specialized with minimal data, offering a pathway to performance that does not rely on massive model scale.

\textit{In sum, few-shot learning leads to more substantial performance improvements in SLMs compared to LLMs, often closing the performance gap or enabling SLMs to surpass their larger counterparts. This highlights the efficiency and adaptability of smaller models, as their performance gains contrast with the more variable and sometimes negligible improvements seen in LLMs. This suggests that SLMs can be effectively specialized for specific mental health screening tasks with minimal data, offering a practical path to high performance without relying on massive model scale.}


\section{Conclusion and Discussions}

SLMs are emerging as a viable, privacy-preserving alternative to LLMs for on-device applications. This paper provides the first systematic evaluation of SLM mental health understanding compared to established LLM baselines. Our findings demonstrate that SLMs achieve competitive performance on binary classification tasks (within 2\% F1 of LLMs), yet both model classes struggle equally with nuanced severity assessment. We also show few-shot learning disproportionately benefits SLMs, making them highly effective for initial screening without extensive fine-tuning. This work validates the potential of SLMs for privacy-conscious mental health applications while highlighting the current limitations of automated clinical evaluation.

Our study is subject to several limitations. First, our evaluation is confined to English-language datasets, primarily from Reddit. The models' performance may not generalize to other languages, cultural contexts, or data sources like clinical notes or transcripts. Second, while in-context learning is a powerful technique, it may not be sufficient for highly complex clinical nuances, where fine-tuning might be required. Lastly, this study focuses on text-based analysis and does not incorporate other data modalities like speech or behavior, which could provide a more holistic view of an individual's mental state. Future work should aim to address these gaps by exploring multilingual models, fine-tuning strategies, and multimodal data fusion.

\vspace{-0.5em}
\section*{Ethical Impact Statement}

This research involves public yet sensitive mental health data and carries significant ethical responsibilities. Our investigation of open sourced SLM understanding capabilities is motivated by the potential for privacy-preserving mental health tools. We recognize the critical importance of safeguarding user privacy and ensuring that data protection protocols meet or exceed current ethical standards.

However, we emphasize that SLMs are designed to supplement and support mental health professionals, rather than replace them. The complexity and nuance of mental health diagnosis and treatment require the expertise of trained clinicians, and AI-driven models should be viewed as tools to enhance accessibility and early detection, not as standalone solutions.
Any deployment must include clear limitations disclosure, user consent and data protection remain paramount, and understanding capabilities require ongoing validation in real-world contexts.


\end{document}